\documentclass[10pt,twocolumn,letterpaper]{article}

\usepackage{wacv}              

\definecolor{wacvblue}{rgb}{0.21,0.49,0.74}
\usepackage[pagebackref,breaklinks,colorlinks,allcolors=wacvblue]{hyperref}
\usepackage{graphicx}
\usepackage{tabularx} 
\usepackage{siunitx}
\usepackage{threeparttable}
\usepackage{multirow}
\usepackage{float}
\usepackage{booktabs}
\def\wacvPaperID{830} 
\def\confName{WACV}
\def\confYear{2026}

\title{UniCalib: Targetless LiDAR-Camera Calibration via Probabilistic Flow on Unified Depth Representations}
\author{Shu Han $^{1}$ \quad
Xubo Zhu $^{2}$ \quad
Ji Wu $^{1}$ \quad
Ximeng Cai $^{1}$ \quad Wen Yang $^{2}$ \quad
Huai Yu $^{2}$ \quad
Gui-Song Xia $^{3}$ \\
{$^{1}$ School of Computer Science, Wuhan University} \quad
{$^{2}$ School of Electronic Information, Wuhan University} \\
{$^{3}$ School of Artificial Intelligence, Wuhan University} \\ 
{\tt\small\{hanshu, zhuxubo, ji.wu, ximengcai, yangwen, yuhuai, guisong.xia\}@whu.edu.cn}
}

\begin{document}
\maketitle
\begin{abstract} 

Online targetless extrinsic LiDAR-camera calibration is essential for robust perception in computer vision applications such as autonomous driving. However, existing methods struggle with the significant modality gap between heterogeneous sensors and fail to handle unreliable correspondences arising from real-world challenges like occlusions and dynamic objects. To address these issues, we introduce UniCalib, a novel method that performs calibration by estimating a probabilistic flow on unified depth representations. UniCalib first bridges the modality gap by converting both the camera images and the sparse LiDAR points into unified, dense depth maps, enabling a unified encoder to learn consistent features. Subsequently, it learns a probabilistic flow field that captures the correspondence uncertainty to improve robustness. This probabilistic approach is reinforced by a reliability map and a perceptually weighted sparse flow loss, which guide the model to suppress the influence of unreliable regions. Experimental results on three datasets validate the accuracy and generalization of UniCalib. In particular, it achieves a mean translation error of $0.550\mathrm{cm}$ and a rotation error of $0.044^\circ$ on the KITTI dataset. \end{abstract}    
\section{Introduction}
\label{sec:intro}
\begin{figure}[ht!]
    \centering
\includegraphics[width=\linewidth]{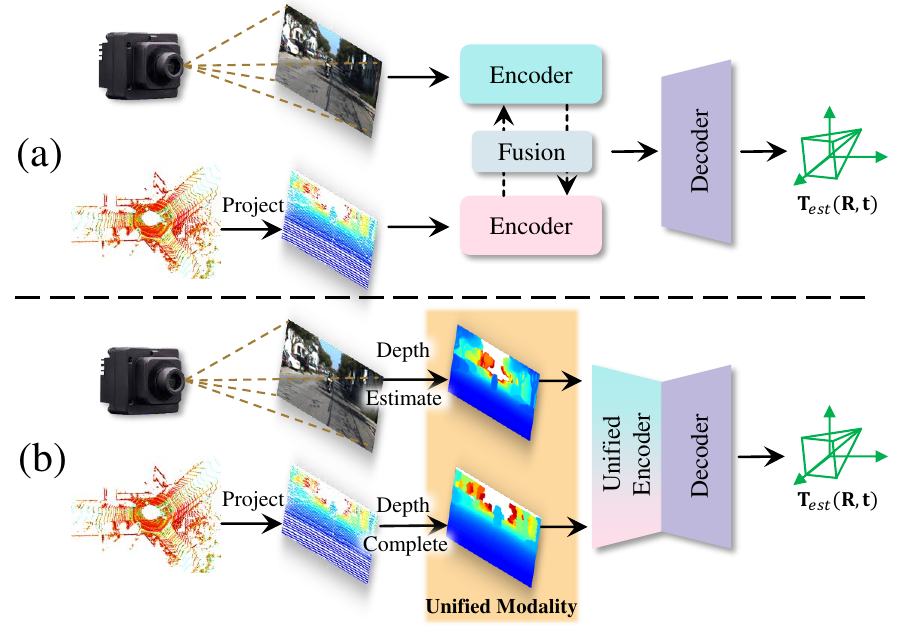}

 \caption{
Architectural comparison for LiDAR-camera calibration. (a) Modality-specific encoders followed by the fusion module. (b) Our method converts both modalities into dense depth maps, enabling a unified encoder.
}

    \label{fig:teaser}
    \vspace{-1em}
\end{figure}
The fusion of camera and Light Detection And Ranging (LiDAR) data is essential for computer vision applications such as autonomous driving and 3D object detection \cite{li2022deepfusion,guo2021liga}, owing to their complementary strengths in capturing texture and geometry.
This requires precise extrinsic calibration to ensure accurate spatial alignment between the sensors. Traditional offline calibration methods \cite{zhang2004extrinsic,unnikrishnan2005fast} rely on handcrafted targets, such as checkerboards. However, in real-world scenarios, mechanical vibrations and operational disturbances cause gradual deviation in the extrinsics, necessitating frequent recalibration. This dependence on specific physical targets is cumbersome and impractical in dynamic environments, making the development of online targetless calibration methods crucial for real-world applications.

Compared to target-based methods, online targetless calibration relies on structured environmental features for registration, requiring consistent and robust cross-modal feature matching between LiDAR and camera data. This is inherently challenging due to the heterogeneous nature of the two modalities. Traditional online targetless methods \cite{castorena2016autocalibration,ma2021crlf,taylor2016motion} bridge the modality gap by leveraging features such as edges, lane markings, and motion trajectories. However, their performance heavily depends on specific scene structures, making them less robust in complex or unstructured environments.

To overcome the reliance on predefined targets, deep learning techniques have been introduced. These methods generally follow two paradigms. End-to-end approaches \cite{xiao2024calibformer,lg} directly regress extrinsics, but their network architectures are coupled with specific camera intrinsics, which limits their generalizability. In contrast, matching-based methods \cite{lv2021cfnet,jing2022dxq} first establish 2D-3D correspondences, decoupling the learning process from camera parameters and thus enhancing generalization. However, as shown in \cref{fig:teaser} (a), a common limitation in both paradigms is the independent extraction of features from the camera image and the LiDAR-projected sparse depth map. Due to inherent modality differences, the extracted features exhibit a significant domain gap. Consequently, these methods struggle with robust cross-modal alignment, which ultimately compromises calibration accuracy. Furthermore, these methods often treat all potential correspondences with uniform importance. This inherent limitation makes them highly vulnerable to real-world challenges, such as dynamic objects and occlusions, as they lack a mechanism to distinguish reliable data from resulting outliers, ultimately degrading calibration accuracy.

In this paper, we propose UniCalib, a novel matching-based method that transforms the cross-modal calibration task into a probabilistic intra-modality depth flow estimation problem. As illustrated in \cref{fig:teaser} (b), we bridge the modality gap by converting both camera and LiDAR data into dense depth maps, thereby unifying them into the same modality. This enables a unified encoder to learn consistent features from the shared domain. To handle real-world uncertainty, our framework predicts a probabilistic depth flow field representing a per-pixel distribution over correspondences, guided by a reliability map that prioritizes valid LiDAR data. This probabilistic approach is reinforced by our novel Perceptually Weighted Sparse Flow (PWSF) loss, which utilizes the predicted uncertainty and an outlier probability map to suppress supervision in unreliable regions (dynamic objects and occlusions), thereby preventing overfitting adaptively. Finally, the robust 2D-3D correspondences derived from this process are fed into a PnP solver to compute the extrinsics.

Our main contributions are listed as follows:
\begin{itemize}
    \item We propose UniCalib, a novel targetless LiDAR-camera calibration method that reframes the 2D-3D correspondence estimation task as a probabilistic intra-modality depth flow estimation problem, addressing the cross-modal challenge. 
    \item  We convert both camera images and sparse LiDAR point clouds into the same modality, which enables a unified feature encoder to learn consistent representations, effectively bridging the modality gap.
    \item We introduce a probabilistic depth flow network featuring a reliability map and a novel PWSF loss, which handles correspondence uncertainty and adaptively suppresses supervision in unreliable regions.  
    \item The experimental results across multiple datasets, including KITTI Odometry, KITTI Raw, and KITTI-360, demonstrate that UniCalib achieves accurate extrinsic calibration results and exhibits reliable generalization ability.
\end{itemize}


\begin{figure*}[!h]
    \centering
    \setlength{\abovecaptionskip}{0pt}
    \includegraphics[width=\textwidth]{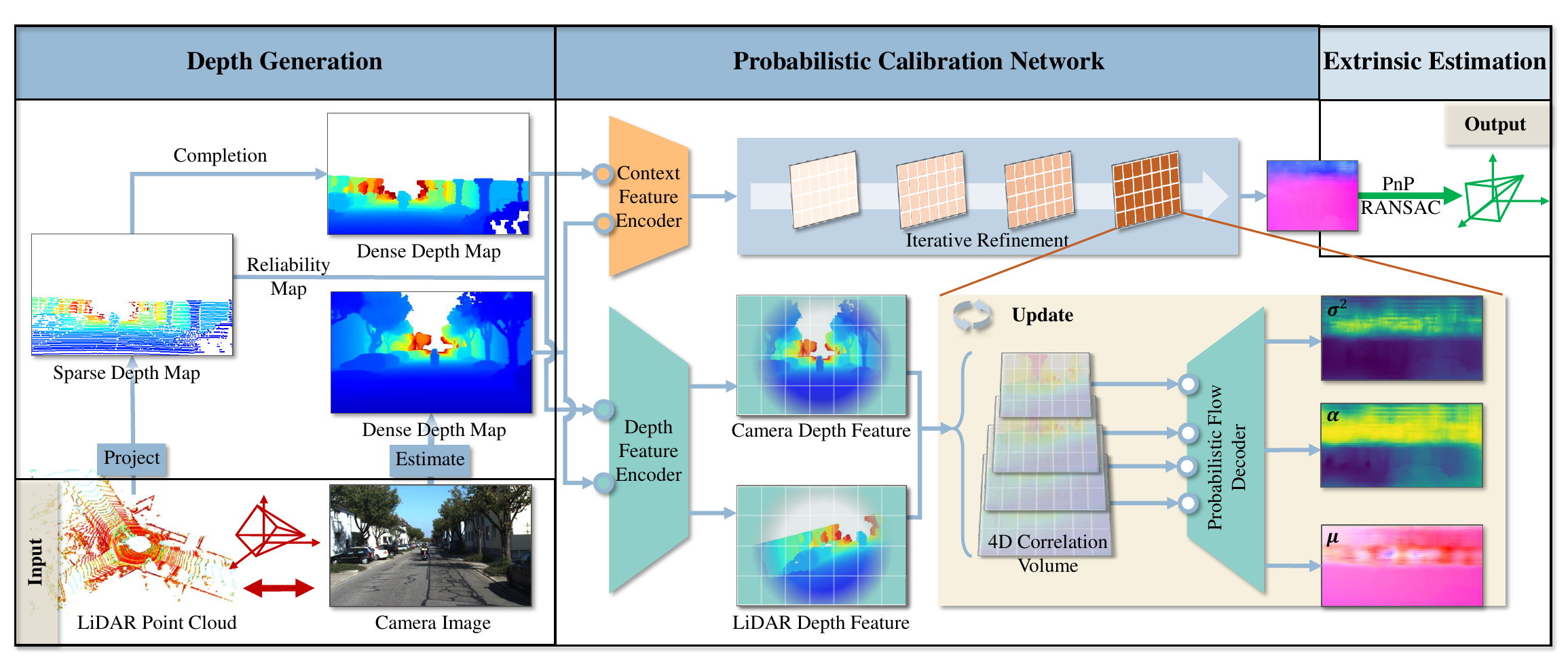}

\caption{
Overview of UniCalib. Depth Generation converts camera and LiDAR inputs into dense depth maps, which are passed to the Probabilistic Calibration Network to estimate a probabilistic depth flow field. The mean of the predicted distribution, referred to as the estimated flow, is then used by the Extrinsic Estimation module to compute the LiDAR-camera extrinsics.
}

    \label{fig:overview}
    \vspace{-18pt}
\end{figure*}
\section{Related Work}
\label{relawork}
LiDAR-camera extrinsic calibration can be broadly classified into target-based and targetless methods. While learning-based approaches are gaining prominence for their calibration accuracy and attracting considerable attention, we consider them a distinct category for further discussion.

\subsection{Target-based Methods}
Target-based LiDAR-camera calibration relies on calibration targets detectable in both modalities to establish accurate 2D-3D correspondences. Early methods \cite{zhang2004extrinsic, unnikrishnan2005fast,Guan_TCYB} are inspired by camera intrinsic calibration using chessboard patterns. Over time, various calibration targets, such as plane boards with holes \cite{yan2023joint}, triangles \cite{debattisti2013automated}, polygons \cite{liao2018extrinsic}, spheres \cite{zhang2024automatic}, and ArUco \cite{dhall2017lidar}, have been developed to enhance feature extraction accuracy. While these methods improve calibration precision, the complex target designs increase production difficulty. Despite the high precision achievable, the reliance on targets introduces limitations related to accessibility and automation, making it unsuitable for online calibration.

\subsection{Targetless Methods}
Targetless methods eliminate the need for calibration targets by leveraging special features and generally establishing correspondences directly from natural scene structures. Frequently used features include edges \cite{castorena2016autocalibration} and lane markings \cite{ma2021crlf}, with extrinsics optimized to maximize feature consistency.

To avoid relying on predefined features, Nagy \textit{et al.} \cite{nagy2019sfm} recover 3D geometry using Structure-from-Motion (SfM) and jointly optimize the extrinsics during the reconstruction process. Mutual information techniques \cite{pandey2012automatic,Guan_IJCV} align projected depth or intensity maps with camera images, optimizing the extrinsics by maximizing mutual information. Taylor and Nieto \cite{taylor2016motion} employ motion-based approaches by matching trajectories but require diverse motion data for accurate optimization.

In summary, targetless methods use a range of features and metrics to formulate an optimization problem for the extrinsics. However, their effectiveness is highly dependent on initial conditions, assumption validity, and scene suitability.

\subsection{Learning-based Methods}

Learning-based methods leverage neural networks to autonomously extract features, eliminating the need for manually designed features and enabling more flexible calibration. These methods can be categorized into two types: regression-based and matching-based.

Regression-based methods employ an end-to-end approach where the network directly regresses extrinsics. RegNet \cite{schneider2017regnet} pioneers this framework, defining a pipeline of feature extraction, matching, and global regression. CalibNet \cite{iyer2018calibnet} further introduces geometric constraints to improve consistency. CalibRCNN \cite{shi2020calibrcnn} and CALNet \cite{shang2022calnet} refine these techniques by incorporating temporal losses for improved stability. CalibDepth \cite{zhu2023calibdepth} enhances feature matching between the sparse LiDAR-projected depth map and the estimated dense depth map but struggles to align complex geometric structures, reducing calibration accuracy. Despite their effectiveness, regression-based methods strongly depend on camera configurations, requiring retraining or fine-tuning for new sensor setups. 

Matching-based methods, in contrast, decouple feature matching from extrinsic estimation, establishing 2D-3D correspondences and formulating calibration as a PnP problem. This improves generalization across different camera configurations. CFNet \cite{lv2021cfnet} introduces calibration flow, leveraging the EPnP algorithm \cite{lepetit2009ep}, while DXQ-Net \cite{jing2022dxq} integrates a differentiable pose estimation module with uncertainty filtering. However, both regression-based and matching-based approaches face fundamental limitations. They struggle to extract consistent features across the modality gap and lack robust mechanisms to handle real-world outliers from sources like dynamic objects or occlusions, leaving them susceptible to spurious correspondences.

Diverging from conventional methods, our approach recasts the cross-modal problem into an intra-modality task by converting both camera images and LiDAR point clouds into dense depth maps. We then estimate a probabilistic depth flow between these maps, utilizing our innovative PWSF loss. This loss function incorporates predicted uncertainty and an outlier probability map to minimize the impact of unreliable areas. This process yields reliable 2D-3D correspondences, from which a PnP solver computes the final extrinsics.

\section{Proposed Method}
\label{sec3}

An overview of our method is shown in \cref{fig:overview}. The inputs are a camera image and a LiDAR point cloud from the same scene. The camera image is processed by MoGe \cite{moge} to estimate a dense depth map, while the LiDAR point cloud is projected into a sparse depth map and subsequently completed to a dense one. These two dense depth maps are then fed into our probabilistic calibration network, with the sparse LiDAR projection provided as supplementary input. The network estimates a distribution over depth flow field to model the 2D–3D correspondences between the camera and LiDAR data. Finally, the extrinsics are estimated using a PnP solver.
 
\subsection{Problem Formulation}
\label{Problem Formulation}

Given the misaligned camera image $I\in\mathbb{R}^{3\times W\times H}$ and LiDAR point cloud $P_L\in\mathbb{R}^{3\times N}$, the extrinsic calibration aims to estimate a 6-DoF extrinsic parameter $\mathbf{T}_{est}=\begin{bmatrix}\mathbf{R}_{est}&\mathbf{t}_{est}\\\mathbf{0}^T&1\end{bmatrix}\in SE(3)$ that adjusts the initial extrinsics $\mathbf{T}_{init}\in SE(3)$ for accurate sensor alignment. Here, $W$ and $H$ represent the image width and height, respectively. $\mathbf{R}_{est}\in SO(3)$ denotes the rotation matrix, and $\mathbf{t}_{est}\in\mathbb{R}^{3\times1}$ represents the translation vector.
For matching-based methods, the 2D-3D correspondences between image pixels and LiDAR points are first established. The extrinsic parameter is then estimated via the PnP solver, with RANSAC to prune outliers and enhance robustness.

\begin{figure}[t!]
    \centering
    \setlength{\abovecaptionskip}{0pt}
    \includegraphics[width=\linewidth]{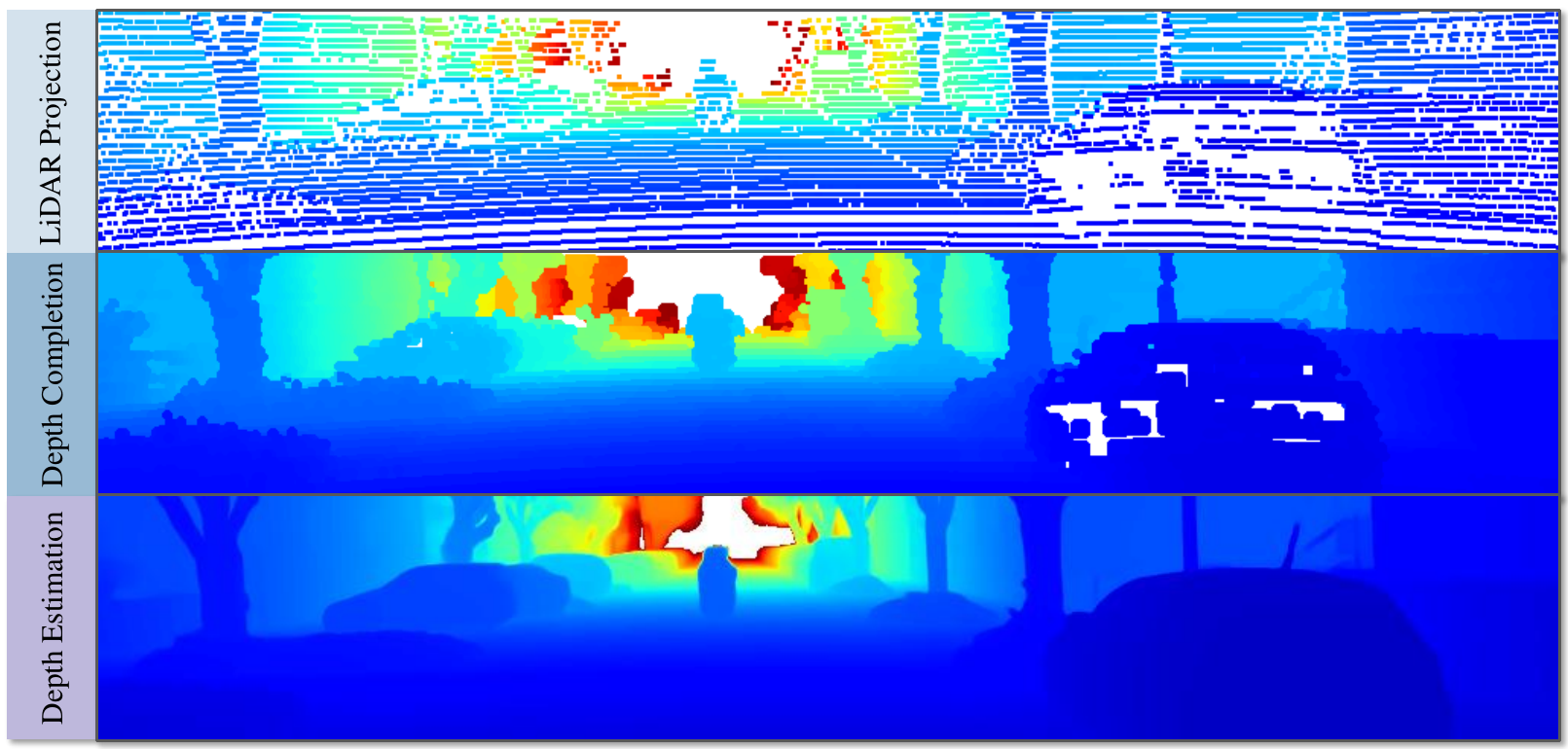}

    \caption{Comparison of different depth maps. After completion, the depth maps generated by the LiDAR and the camera exhibit similar textures.}
    \label{fig:depth_compare}
    \vspace{-1em}
\end{figure}
\subsection{Depth Generation}
\label{Depth Generation}
The main challenge in LiDAR-camera extrinsic calibration lies in cross-modal data matching due to structure and information density differences between point clouds and images. To address this, we propose a novel method that transforms both modalities into dense depth maps, aligning their structure and information density for effective matching. First, we employ MoGe \cite{moge} to estimate the dense depth map $D_I^d\in\mathbb{R}^{1\times W\times H}$ from a camera image. Meanwhile, leveraging the camera’s intrinsic parameter $\mathbf{K}\in\mathbb{R}^{3\times3}$ and initial extrinsic parameter 
$\mathbf{T}_{init}=\begin{bmatrix}\mathbf{R}_{init}&\mathbf{t}_{init}\\\mathbf{0}^T&1\end{bmatrix}\in SE(3)$, the LiDAR point clouds $P_L$ are projected onto the pixel coordinate system, resulting in a sparse depth map $D_L^s\in\mathbb{R}^{1\times W\times H}$. The projection process can be formulated as
{\setlength{\abovedisplayskip}{0.5em}
\setlength{\belowdisplayskip}{0.5em}
\begin{equation}
\scalebox{0.80}{$
d_i\begin{bmatrix}u_i \\ v_i \\ 1\end{bmatrix} = 
\mathbf{K}\left(\mathbf{T}_{init}\begin{bmatrix}x_i \\ y_i \\ z_i \\ 1\end{bmatrix}\right)_{(1 \sim 3)} =
\mathbf{K}\left(\mathbf{R}_{init}\begin{bmatrix}x_i \\ y_i \\ z_i\end{bmatrix} + \mathbf{t}_{init}\right)
$},
\end{equation}
}%
where $d_{i}$ represents the depth of each pixel, and $[x_i,y_i,z_i,1]^\mathrm{T}$ is the homogeneous coordinate of the 3D point, while $[u_i,v_i,1]^\mathrm{T}$ is the corresponding pixel's homogeneous coordinate in 2D. However, due to the inherent sparsity of LiDAR data, $D_L^s$ contains only a limited number of valid pixels, leading to a significant information density disparity between $D_L^s$ and $D_I^d$, which adversely affects subsequent depth flow estimation. To mitigate this issue, we employ an efficient depth completion method \cite{dilation} to transform $D_L^s$ into a dense depth map $D_L^d\in\mathbb{R}^{1\times W\times H}$. As shown in \cref{fig:depth_compare}, this process ensures that both the LiDAR point cloud and the camera image are effectively converted into dense depth maps, facilitating the subsequent depth flow estimation.

Inspired by optical flow, we define depth flow $\mathbf{f} \in \mathbb{R}^{2 \times W \times H}$ as a pixel-wise displacement between two depth maps, capturing horizontal and vertical shifts. It encodes dense pixel correspondences and establishes 2D-3D correspondences between the camera and LiDAR for extrinsic estimation.

To supervise the depth flow, we calculate the ground truth depth flow $\mathbf{f}_{gt}$ based on the known discrepancy between the initial extrinsic parameter $\mathbf{T}_{init}$ and the ground truth $\mathbf{T}_{gt}$. Given this, the pixel location $p_i = [u_i, v_i]^T \in \mathbb{R}^2$ in $D_L^s$ deviates from its ground truth position $p_i^{gt} = [u_i^{gt}, v_i^{gt}]^T \in \mathbb{R}^2$. The ground truth depth flow $\mathbf{f}_{gt}$, which captures this displacement, is computed as
{\setlength{\abovedisplayskip}{0.5em}
\setlength{\belowdisplayskip}{0.5em}
\begin{equation}
\mathbf{f}_{gt}(p_i) = p_i^{gt} - p_i.
\end{equation}  
}%
Here, $\mathbf{f}_{gt}$ is further utilized for loss computation to enhance the calibration accuracy.

\subsection{Probabilistic Calibration Network}
\label{network}

Our network architecture, inspired by SEA-RAFT~\cite{searaft}, integrates LiDAR-projected and camera-predicted depth maps into a unified probabilistic framework for depth flow inference.

\subsubsection{Depth Feature Encoder}
\label{feature encoder}
As $D_L^d$ incorporates richer geometric structures, it becomes more texturally similar to the dense camera depth map $D_I^d$, as shown in \cref{fig:depth_compare}. To exploit this enhanced similarity, we normalize both depth maps to the range $[-1,1]$ and feed them into a unified ResNet-34 depth encoder. This enables consistent, modality-invariant feature extraction, improves training efficiency through parameter sharing, and helps the network learn a shared feature space, improving cross-modal calibration accuracy. Additionally, a separate ResNet-34 context encoder extracts context features $F_C$ and the initial flow $\mathbf{f}_{init}$ from the normalized depth maps

However, depth completion introduces noise and blur, leading to unreliable features and degraded calibration accuracy. To address this issue, we predict a reliability map $R$ from the sparse depth $D_L^s$ via a depth encoder, guiding feature extraction by emphasizing reliable pixels. Specifically, we perform element-wise multiplication between the features $F_L$ extracted from the completed depth $D_L^{d}$ and $R$ to obtain reliability-weighted features $F_L^R$, which enhances the accuracy of feature correlation.

Following RAFT~\cite{teed2020raft}, we construct a multi-scale 4D correlation volume by computing all-pairs dot products between the depth features $F_I$ extracted from $D_I^{d}$ and the downsampled $F_L^R$ at each level $k$:
{\setlength{\abovedisplayskip}{3pt}
 \setlength{\belowdisplayskip}{3pt}
\begin{equation}
V_{k} = F_I \circ \texttt{AvgPool}(F_L^R, 2^k)^\top \in \mathbb{R}^{h\times w\times \frac{h}{2^k} \times \frac{w}{2^k}},
\end{equation}
}

where $\circ$ denotes the dot product between feature vectors. We set $(h, w) = \frac{1}{8}(H, W)$ and use $k = 4$ to balance efficiency and accuracy. The resulting correlation volume provides rich similarity cues across scales, serving as the input to the subsequent probabilistic flow decoder.
\subsubsection{Probabilistic Flow Decoder}
\label{sec:decoder}
To address the ambiguity in depth flow estimation caused by occlusions and dynamic objects, we design a probabilistic decoder that jointly predicts the depth flow and uncertainty, allowing the model to express confidence and improve robustness in challenging regions. 
Given a pair of depth maps $X=\left(D_I^d, D_L^d \right)$, the goal is to estimate a flow $\mathbf{f} \in \mathbb{R}^{2 \times W \times H}$ that warps $D_L^d$ to align with $D_I^d$. Instead of directly regressing $\mathbf{f}$ via a network $F$ with parameters $\theta$, we model its conditional probability distribution $p(\mathbf{f}|X; \theta)$. To achieve this, the decoder predicts the parameters $\Phi(X; \theta)$ of a family of distributions $p(\mathbf{f}| X; \theta) = p(\mathbf{f}|\Phi(X; \theta)) = \prod_{uv} p(f_{uv} | \varphi_{uv}(X; \theta))$.  Here, $f_{uv} \in \mathbb{R}^2$ is the flow at pixel $(u,v)$, while $\varphi_{uv} \in \mathbb{R}^n$ are the corresponding predicted parameters. For simplicity, we omit the subscript $uv$ in what follows. In this case, the density is represented using Laplace distributions as follows:

\begin{equation}
\label{eq:laplace}
p(f| \mu, \sigma^2) =\frac{1}{\sqrt{2 \sigma_u^2}} e^{-\sqrt{\frac{2}{\sigma_u^2}}|u-\mu_u|} . \frac{1}{\sqrt{2 \sigma_v^2}}
e^{-\sqrt{\frac{2}{\sigma_v^2}}|v-\mu_v|},
\end{equation}
where $f = (u,v) \in \mathbb{R}^2$ denotes the flow vector with conditionally independent components. The network predicts the mean $\mu = [\mu_u, \mu_v]^T$ as the flow and a shared variance $\sigma^2$ to represent uncertainty, such that $\sigma_u^2 = \sigma_v^2 = \sigma^2$ for efficiency and stability.

The decoder follows an iterative refinement strategy inspired by RAFT~\cite{teed2020raft}, but differs in that our output is four-dimensional. Specifically, it predicts distribution parameters $\Phi$ consisting of a two-dimensional mean $\boldsymbol{\mu}$ and a one-dimensional variance $\boldsymbol{\sigma}^2$, as well as a one-dimensional outlier probability map $\boldsymbol{\alpha}$. The update block takes the current flow estimate, the 4D correlation volume, and context features $F_C$ as input at each iteration. The final optimized mean $\boldsymbol{\mu}$ is treated as the predicted flow and passed to a PnP solver to estimate the extrinsics. Both the predicted variance $\boldsymbol{\sigma}^2$ and outlier probability $\boldsymbol{\alpha}$ are used in the loss function to supervise the distribution, as detailed in \cref{loss}.


\subsection{Training Objective}
\label{loss}
\subsubsection{Valid Mask}
As $\mathbf{f}_{gt}$ is derived from the sparse LiDAR projection, it is only valid at limited pixels. Applying loss over all pixels would introduce noise from invalid regions and degrade depth flow estimation. To mitigate this, we introduce a valid mask $v(p)$ to filter out invalid pixels, defined as:
\begin{equation}
v(p) = 
\begin{cases}
1, & \mathbf{f}_{gt}(p) \neq 0, \\
0, & otherwise.
\end{cases}
\end{equation}

\subsubsection{Negative Log-likelihood Loss}

Previous probabilistic regression methods~\cite{chen2022aspanformer,sun2021loftr} often employ the negative log-likelihood (NLL) loss, which, for the Laplace distribution, is defined as

\vspace{-1em}
\begin{align}
\mathcal{L}(\boldsymbol{\mu_{gt}}, \boldsymbol{\mu}, \boldsymbol{\sigma}^2) 
= &\frac{1}{HW} \sum_{p} \Bigg[
 \log \left( \frac{1}{2\sigma^2(p)} \right) \nonumber \\
& + \sqrt{ \frac{2}{\sigma^2(p)} } 
\left\| \boldsymbol{\mu_{gt}}(p) - \boldsymbol{\mu}(p) \right\|_1
\Bigg],
\end{align}
where $\boldsymbol{\mu}$ and $\boldsymbol{\mu}_{gt}$ are the predicted and ground truth flows, and $\boldsymbol{\sigma}^2$ is the predicted per-pixel variance, which modulates confidence by downweighting high-uncertainty errors. However, this formulation is not well-suited for flow estimation tasks\cite{zhang2023rgm}, especially depth flow, as it exhibits two inherent limitations: the application of uniform supervision irrespective of prediction confidence, and the tendency to overfit uncertain regions by inflating the predicted variance.

\subsubsection{Perceptually Weighted Sparse Flow Loss}

To address this, we propose the Perceptually Weighted Sparse Flow (PWSF) loss, which leverages the model's estimated confidence to adaptively suppress the influence of outliers during supervision. Specifically, the loss is weighted by a predicted outlier probability map $\boldsymbol{\alpha}\in [0,1]^{H\times W}$, which down-weights supervision in regions likely to contain mismatches or occlusions. The PWSF loss comprises two complementary components, defined as:
\begin{align}
\mathcal{L}_{\mathrm{PWSF}} 
= \frac{1}{N_{valid}} \sum_{p} v(p) \Big[ 
& (1 - \alpha(p))\, \mathcal{L}(\mathbf{f}_{gt}, \boldsymbol{\mu}, \mathbf{2}) \nonumber \\
& + \alpha(p)\, \mathcal{L}(\mathbf{f}_{gt}, \boldsymbol{\mu}, \boldsymbol{\sigma}^2)
\Big],
\vspace{-1em}
\end{align}
where $N_{valid} = \sum_p v(p)$ denotes the total number of valid pixels. When $\alpha(p)$ is small, indicating a high likelihood of inliers, the supervision is dominated by a standard L1 loss, which corresponds to the NLL loss of a Laplace distribution with fixed variance $\boldsymbol{\sigma}^2 = \mathbf{2}$. Conversely, when $\alpha(p)$ is large, indicating likely outliers, a variance-weighted formulation takes precedence, allowing the network to attenuate the influence of unreliable predictions via a learned uncertainty map.

The PWSF loss adaptively downweights unreliable predictions while preserving strong supervision for confident ones. It mitigates variance inflation in unreliable regions and maintains overall flow sparsity, leading to improved robustness in real-world scenarios.

\subsubsection{Total Loss}
Finally, the total loss is computed as an exponentially weighted sum of the PWSF loss across all iterations: 
\vspace{-0.7em}
\begin{equation}
    \mathcal{L}_{total}=\sum_{i=1}^N\gamma^{N-i}\mathcal{L}_{PWSF}^i, 
\end{equation}
where $N$ denotes the total number of iterations, and $\gamma=0.8$ is an exponential decay factor that reduces the influence of later iterations. $\mathcal{L}_{PWSF}^i$ is the PWSF loss at the $i$-th iteration. This weighted summation ensures that earlier iterations contribute more to the optimization process, promoting a more stable and progressive refinement.


\vspace{-0.5em}
\begin{table}[t!]

\centering
\caption{Different experimental configurations}
\label{tab:exp_set}
\resizebox{\columnwidth}{!}{
\begin{tabular}{ccccc}
\hline
\textbf{Name} & \textbf{Dataset} & \textbf{Perturbation} & \textbf{Training} & \textbf{Test} \\ \hline
Exp1 & KITTI Odometry & 5\si{\degree}, 0.10\si{\m} & 01-21 & 00 \\ \hline
Exp2 & KITTI Odometry & 10\si{\degree}, 0.25\si{\m} & 01-21 & 00 \\ \hline
Exp3 & KITTI Raw & 10\si{\degree}, 0.25\si{\m} & 2011-9-26 & 2011-9-30 \\ \hline
Exp4 & KITTI-360 & 10\si{\degree}, 0.25\si{\m} & / & Test-SLAM \\ \hline
\end{tabular}
}
\vspace{-1em}
\end{table}

\section{Experiments}
\begin{table*}[tbp]
\belowrulesep=2pt
\aboverulesep=0pt
    \centering

        \caption{Comparison results on different experimental configurations}
        \label{tab:compare}
\resizebox{0.85\textwidth}{!}{%
        \begin{tabular}{cccccccccccccc}
            \toprule
            \multirow{2}[2]{*}{Dataset} & \multirow{2}[2]{*}{Method} & & \multicolumn{4}{c}{Translation (\si{\cm}) $\downarrow$ } & & \multicolumn{4}{c}{Rotation (\si{\degree}) $\downarrow$ } \\
            \cmidrule(lr){4-7} \cmidrule(lr){9-12}
             & & & Mean & X & Y & Z & & Mean & Roll & Pitch & Yaw \\
            \midrule
            \multirow{3}{*}{Exp1} 
            & NetCalib\cite{wu2021netcalib} & & 1.291 & 1.618& 0.917&1.337 & & 0.125 & 0.083 & 0.189 & 0.103 \\
            & DXQ-Net~\cite{jing2022dxq} & & 0.774 & 0.754 & 0.476 & 1.091 & & \textbf{0.042} & 0.049 & 0.046 &\textbf{ 0.032} \\
            & UniCalib & & \textbf{0.550} & \textbf{0.747} & \textbf{0.461} & \textbf{0.444} & & 0.044 & \textbf{0.046} &\textbf{ 0.041} & 0.043 \\
            \midrule
            
            \multirow{4}{*}{Exp2} 

            & CALNet\cite{shang2022calnet} & & 3.03 & 3.65 & 1.63& 3.80 & & 0.20 & 0.10 & 0.38 & 0.12 \\
            & PSNet\cite{wu2022psnet} & & 3.1 & 3.8 & 2.8 & 2.6  & & 0.15 & 0.06 & 0.26 & 0.12 \\
            & CalibFormer\cite{xiao2024calibformer} & & 1.188 & 1.101 & 0.902 & 1.561& & 0.141 & 0.076 & 0.259 & 0.087 \\
            & UniCalib & & \textbf{0.913} & \textbf{1.063} & \textbf{0.803} & \textbf{0.872}& & \textbf{0.053} & \textbf{0.055} & \textbf{0.054} &\textbf{0.051} \\
            \midrule
            \multirow{4}{*}{Exp3}

            & CalibDepth\cite{zhu2023calibdepth}& & 4.75 & 6.66 &\textbf{ 1.12 }& 6.48 & & 0.348 & 0.180 & 0.862 & 0.181 \\
            & NetCalib\cite{wu2021netcalib} & & 4.38 & 6.55 &3.10& 3.50 & & 0.378 & 0.200 & 0.561 & 0.372 \\
            & CalibLG\cite{lg}& & 3.81 & 3.24 &3.93& 4.25 & & 0.240 & 0.086 & 0.361 & 0.280 \\
            & UniCalib & & \textbf{2.368} &\textbf{ 3.400} & 2.259 & \textbf{1.444} & & \textbf{0.084} & \textbf{0.101} & \textbf{0.068} & \textbf{0.082} \\
            \midrule

            \multirow{2}{*}{Exp4} 
            & CalibDepth\cite{zhu2023calibdepth} & & 18.796&26.987 & 12.208 & 17.193  & & 2.418 & 3.833 &  0.752&2.650 \\
            & UniCalib & & \textbf{9.664} & \textbf{10.408} & \textbf{9.299} & \textbf{9.285} & & \textbf{0.381} & \textbf{0.443} &\textbf{ 0.386} & \textbf{0.314} \\
            \bottomrule
        \end{tabular}
        }
    \vspace{-3pt}
\end{table*}

In this section, we evaluate UniCalib on the KITTI Raw \cite{kitti}, KITTI Odometry \cite{kitti}, and KITTI-360 \cite{360} datasets to compare it with state-of-the-art approaches, validating its effectiveness and generalization capability. 
\subsection{Experimental Setup}
\subsubsection{Dataset Preparation}

Following existing methods \cite{xiao2024calibformer,zhu2023calibdepth}, a random perturbation $\Delta\mathbf{T}$ within a certain range is added to the ground truth extrinsic parameter $\mathbf{T}_{gt}$ to obtain the initial extrinsic parameter $\mathbf{T}
_{init}=\Delta\mathbf{T}^{-1}\times\mathbf{T}_{gt}$. The LiDAR point cloud is transformed using $\mathbf{T}_{init}$ to ensure that the updated ground truth extrinsic parameter $\mathbf{T}_{gt}^{'}$ is $\Delta\mathbf{T}$.  

As no standardized setting exists for LiDAR-camera extrinsic calibration, we establish several configurations to facilitate fair comparisons, as presented in \cref{tab:exp_set}.

Exp1 follows the configuration of DXQ-Net \cite{jing2022dxq}, where sequences 01 to 21 of the KITTI Odometry dataset are used for training and validation, and sequence 00 is used for testing, with the initial perturbation of $(5^\circ,0.10\mathrm{m})$.

Exp2 follows the configuration of CalibFormer \cite{xiao2024calibformer}, where sequences 01 to 21 of the KITTI Odometry dataset are used for training and validation, while sequence 00 is used for testing. The initial perturbation is set to $(10^\circ,0.25\mathrm{m})$.

Exp3 is based on CalibDepth \cite{zhu2023calibdepth} and CalibLG \cite{lg}, where 24,000 pairs from sequence 2011-9-26 of the KITTI Raw dataset are used for training and validation, and sequence 2011-9-30 is used for testing. The initial perturbation remained $(10^\circ,0.25\mathrm{m})$.  

Exp4 is designed to evaluate the generalization ability of the proposed method. The model trained using the Exp3 setup is directly tested on the Test-SLAM subset of the KITTI-360 dataset, with the same initial perturbation as Exp3 $(10^\circ,0.25\mathrm{m})$.  

\subsubsection{Implementation Details}
Our network is adapted from SEA-RAFT~\cite{searaft}, with a modified ResNet-34 backbone for depth feature extraction. The probabilistic flow decoder is implemented as described in \cref{sec:decoder}. Data augmentation includes random cropping and horizontal flipping. All images are cropped to a resolution of 960~$\times$~320, and the camera intrinsics are adjusted accordingly to ensure accurate LiDAR point cloud projection under varying image crops. During training, the AdamW optimizer is used with a weight decay of $1\times10^{-5}$ and an initial learning rate of $1\times10^{-4}$. The OneCycleLR scheduler is employed for learning rate adjustment. The model is trained for 100 epochs with a batch size of 6, and the number of iterative refinements is set to 4 during training. All the experiments are conducted on a single NVIDIA RTX 4090 GPU.

\subsubsection{Evaluation Metrics}
The evaluation of LiDAR-camera extrinsic calibration methods is based on the rotation and translation errors between the estimated extrinsic parameter $\mathbf{T}_{est}$ and the ground truth $\mathbf{T}_{gt}^{'}$. The rotation error is measured using Euler angles, where the mean absolute errors are computed separately for roll, pitch, and yaw. The translation error is assessed by calculating the mean absolute distance error along the x, y, and z directions. 
\begin{figure}[!t]
    \centering
    \vspace{-1em}
\includegraphics[width=\columnwidth]{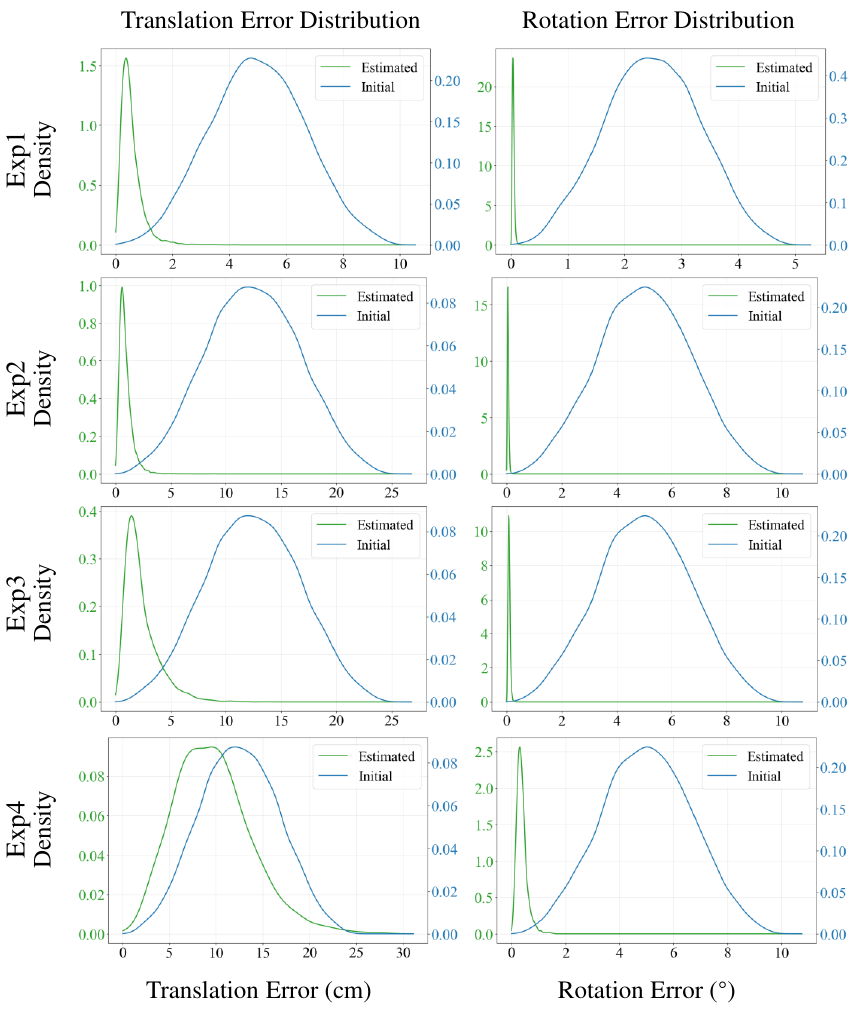}
    \caption{Error distributions under four different experimental configurations (Exp1-Exp4). 
    }
    \label{fig:curves}
    \vspace{-1em}
\end{figure}

\subsection{Experimental Results}

In this section, we analyze the quantitative and qualitative results of our method compared to other methods. Additionally, we examine its runtime performance.

\subsubsection{Quantitative Results} 
As shown in \cref{tab:compare}, UniCalib is evaluated against seven state-of-the-art approaches across various initial deviations and datasets, including NetCalib \cite{wu2021netcalib}, DXQ-Net \cite{jing2022dxq}, CALNet \cite{shang2022calnet}, PSNet \cite{wu2022psnet}, CalibFormer \cite{xiao2024calibformer}, CalibDepth \cite{zhu2023calibdepth} and CalibLG \cite{lg}. \cref{fig:curves} illustrates the translation and rotation error distributions from four distinct experiments, confirming that our method consistently achieves high-accuracy calibration results across all tested scenarios.

\textbf{Exp1:}
UniCalib achieves a mean translation error of $0.550\mathrm{cm}$ and a mean rotation error of $0.044^\circ$. When compared to DXQ-Net, a matching-based method, UniCalib maintains comparable rotation error while greatly reducing the translation error by $28.9\%$. This improvement stems from the PWSF loss, which suppresses unreliable gradients from uncertain regions, leading to more robust depth flow estimation and improved calibration accuracy.

\begin{figure*}[!htp]

    \centering
    \setlength{\abovecaptionskip}{2pt}
    \includegraphics[width=\textwidth]{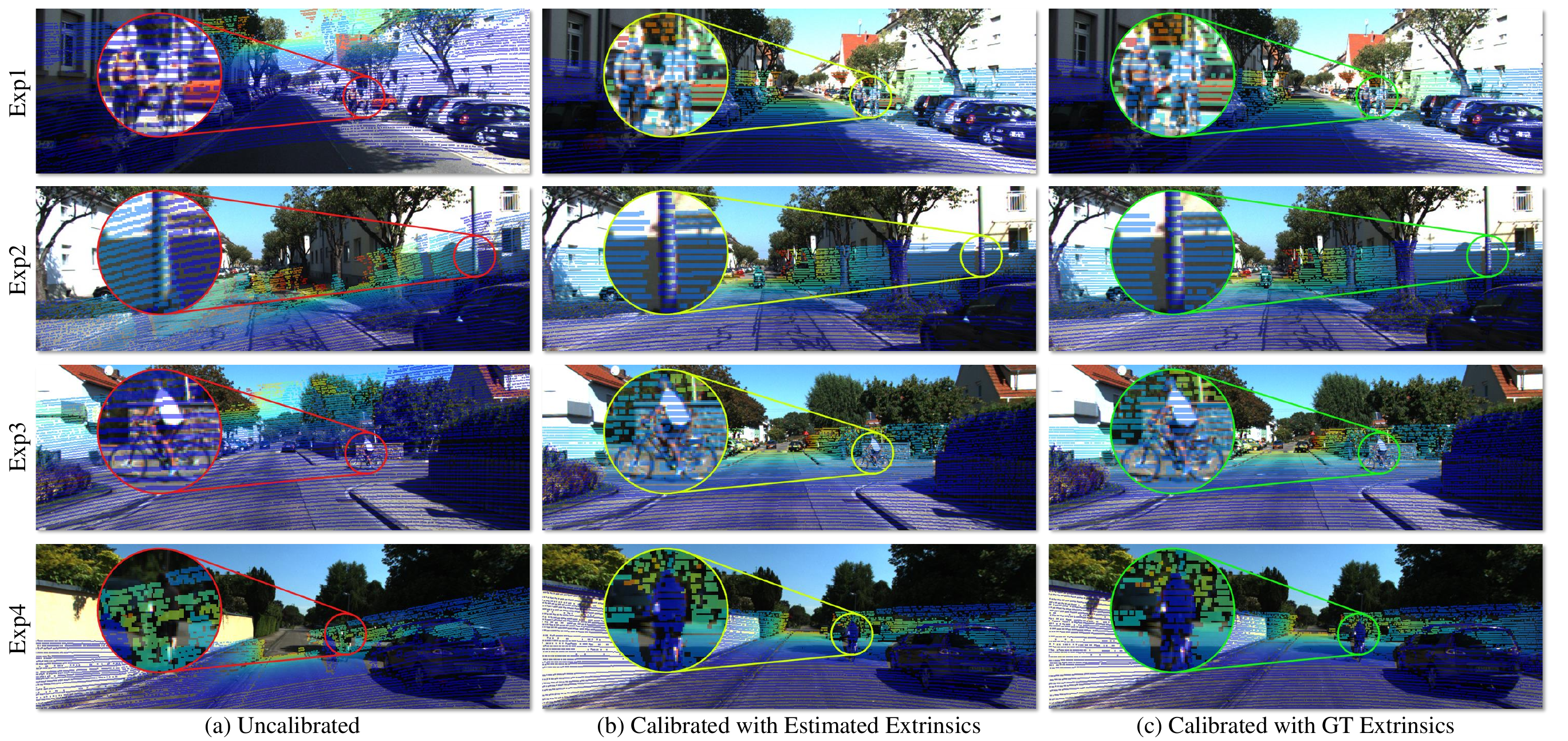}
    
    \caption{Comparison of LiDAR projection on RGB images under four different experimental configurations (Exp1-Exp4). Each column shows the result with: (a) initial uncalibrated extrinsics, (b) extrinsics estimated by UniCalib, and (c) GT extrinsics.
    }
    \label{fig:qualitative}
\end{figure*}

\textbf{Exp2$\&$3:}  
UniCalib achieves a mean translation error of $0.913\mathrm{cm}$ and a mean rotation error of $0.053^\circ$ in Exp2, surpassing CalibFormer, which regresses the extrinsics from a camera image and a sparse LiDAR depth map, by $23.1\%$ and $62.4\%$, respectively. In Exp3, it attains a mean translation error of $2.368\mathrm{cm}$ and a mean rotation error of $0.084^\circ$, outperforming CalibLG, another regression-based method, by $37.8\%$ and $65.0\%$. These improvements arise from our intra-modality formulation, which converts inputs to a unified depth representation, thereby ensuring consistent feature extraction for calibration and significantly boosting the accuracy of both translation and rotation estimation.

\textbf{Exp4:}  
UniCalib achieves a mean translation error of $9.664\mathrm{cm}$ and a mean rotation error of $0.381^\circ$ on the KITTI-360 dataset without being trained on it, outperforming CalibDepth, the only open-source method among the baselines, by $48.6\%$ and $84.2\%$, respectively. This strong generalization ability stems from our probabilistic matching within a unified depth representation, ensuring robustness against both visual appearance variations and geometric outliers.

\subsubsection{Qualitative Results} As shown in \cref{fig:qualitative}, our estimated extrinsics closely aligns with the ground truth across diverse scenarios and initial deviation conditions, highlighting its robustness to a wide range of initial errors and effectiveness in handling complex environments.

\begin{table}
  \centering
  \caption{Ablation studies on different components in UniCalib}
  \label{tab:ablation}
  \begin{threeparttable}
    \resizebox{\columnwidth}{!}{%
    \begin{tabular}{cccccc}
      \toprule
      \multirow{2}{*}{Method} & \multicolumn{2}{c}{KITTI Odometry} & \multicolumn{2}{c}{KITTI Raw}  \\ 
      \cmidrule(lr){2-3} \cmidrule(lr){4-5} 
      & Trans. (\si{\cm}) $\downarrow$  & Rot. (\si{\degree}) $\downarrow$ & Trans. (\si{\cm}) $\downarrow$ & Rot. (\si{\degree})  $\downarrow$ \\ 
      \midrule
      w/o D& 3.673 & 1.563 & 6.842 & 3.122   \\
      w/o U & 1.001 & 0.058 & 2.452 & 0.093   \\
      w/o D, U & 1.141 & 0.062 & 2.92 & 0.109   \\
      w/o R& 1.048 & 0.062 & 2.73 & 0.100   \\
      w/o P (L1)& 1.123& 0.066 & 2.911 & 0.104\\
      w/o P (NLL)& 1.095& 0.065 & 2.85 & 0.102\\
      
      w/ all & \textbf{0.913} & \textbf{0.053} & \textbf{2.368} & \textbf{0.084}  \\
      \bottomrule
    \end{tabular}
    }
    \begin{tablenotes}[para,flushleft]
  \footnotesize
  \begin{minipage}{\columnwidth}
D, U, R, and P denote Completed Dense Depth Map, Unified Feature Encoder, Reliability Map, and PWSF Loss, respectively. “w/o P (L1)” and “w/o P (NLL)” denote replacing PWSF loss with L1 and NLL losses, respectively.
  \end{minipage}
\end{tablenotes}
  \end{threeparttable}
\vspace{-1em}
\end{table}

\subsubsection{Ablation Study}
We conduct ablation experiments on the KITTI Odometry and KITTI Raw datasets under an initial perturbation of $(10^\circ, 0.25 \mathrm{m})$ to evaluate the effectiveness of our proposed components, using mean absolute translation error and mean absolute rotation error as evaluation metrics. The results are shown in \cref{tab:ablation}.  

Without completing the sparse LiDAR depth map and directly using it with the image-estimated dense depth map $D_I^d$ via a unified feature encoder, the error approximately tripled, demonstrating that depth completion reduces the information density gap and enhances feature consistency. When processing $D_L^d$ and $D_I^d$ with separate encoders instead of a unified one, the error increased by about $10\%$, and model parameters grew from 19M to 28M, highlighting the unified encoder’s efficiency in feature extraction and network complexity reduction. Removing both depth completion and the unified encoder led to an approximate $25\%$ increase in error and a $47\%$ rise in model parameters, reinforcing their importance.  

Removing the constraint of the Reliability Map on dense depth features resulted in an error increase of around $15\%$ on both datasets, indicating its role in suppressing noise introduced by depth completion. Replacing the PWSF loss with a standard $\mathcal{L}_1$ loss increases the error by $23\%$, while using an NLL loss leads to a $19\%$ degradation. These results demonstrate that our PWSF loss provides more effective uncertainty-aware supervision by better capturing spatial reliability and suppressing unreliable regions.

\begin{table}[t!]
\centering
\caption{\textsc{Runtime Analysis}}
\label{tab:runtime}
\resizebox{\linewidth}{!}{%
\begin{tabular}{cccc|c} 
    \toprule
    & Depth Generation & Network Inference & PnP Solver & Total \\ 
    \midrule
    Time [ms]   & 37  & 29 & 25 & 91($\sim$11Hz)  \\
    \bottomrule
\end{tabular}%
}
\end{table}

\subsubsection{Runtime Analysis} 
To evaluate the computational efficiency of our method, we perform a comprehensive runtime analysis, with the result presented in \cref{tab:runtime}. The initial depth generation stage consists of two concurrent modules: depth estimation of camera image(37 ms) and LiDAR projection with depth completion (12 ms). As these operate in parallel, the latency of this stage is bottlenecked by the depth estimation process. Following this, the network inference and the PnP solver consume 29 ms and 25 ms, respectively. With an input resolution of 960x320, the total pipeline latency is approximately 91 ms, enabling a calibration throughput of about 11 Hz. This demonstrates that UniCalib achieves real-time performance, making it suitable for applications in practical scenarios such as autonomous driving.
\section{Conclusion}
\label{sec5}
In this paper, we presented UniCalib, a novel method that reformulates the cross-modal calibration task as a probabilistic intra-modality depth flow estimation problem. UniCalib bridges the modality gap by generating dense depth maps from both camera and LiDAR inputs, enabling a unified encoder to learn consistent features within a shared representation space. To address real-world uncertainty, our framework predicts a probabilistic depth flow field, guided by a reliability map and enhanced by the PWSF loss that adaptively suppresses supervision in unreliable regions. Experimental results on multiple datasets demonstrate that UniCalib achieves accurate extrinsic calibration with strong generalization capability. In future work, we plan to extend this probabilistic flow estimation framework to other cross-modal alignment problems, such as radar-camera calibration, where handling sparse and noisy measurements remains a key challenge.

\clearpage

{
    \small
    \bibliographystyle{ieeenat_fullname}
    \bibliography{ref.bib}
}

\end{document}